\documentclass{article}

\usepackage[utf8]{inputenc} 
\usepackage[T1]{fontenc}    
\usepackage{hyperref}       
\usepackage{url}            
\usepackage{svg}
\usepackage{booktabs}
\usepackage{amsfonts}       
\usepackage{nicefrac}       
\usepackage{microtype}      
\usepackage{hyperref}
\usepackage{graphicx}
\usepackage{xcolor}
\usepackage{amsmath}
\usepackage{soul}
\usepackage{authblk}

\title{Interactive Latent Interpolation on MNIST Dataset}

\author[1]{Mazeyar Moeini Feizabadi}
\author[2]{Ali Mohammed Shujjat}
\author[3]{Sarah Shahid }
\author[4]{Zainab Hasnain}
\affil[1,2,3,4]{Habib University}

\begin{document}

\maketitle

\begin{abstract}

This paper will discuss the potential of dimensionality reduction with a web-based use of GANs. Throughout a variety of experiments, we show synthesizing visually-appealing samples, interpolating meaningfully between samples, and performing linear arithmetic with latent vectors. GANs have proved to be a remarkable technique to produce computer generated images, very similar to an original image. This is primarily useful when coupled with dimensionality reduction as a effective application of our algorithm. We proposed a new architecture for GANs, which ended up not working for mathematical reasons later explained. We then proposed a new web-based GAN that still takes advantage of dimensionality reduction to speed generation in the browser to .2 milliseconds. Lastly, we made a modern UI with linear interpolation to present the work. With the speedy generation we can generate so fast that we can create a animation type effect that we have never seen before that works on both \textcolor{blue}{\href{https://mazy1998.github.io/browserGAN/}{web and mobile}}.

\end{abstract}

\section{Introduction}

The interactive display of latent interpolation on MNIST dataset is done using GANs. Generative Adversarial Networks, as its name suggests, work like on the basis of two adversarial frameworks. The first is the generative model which learns from the original datasets and tries to copy it. It learns from the input images in the process of producing 'fake' images. \newline \newline
On the other had, the discriminative model tries to distinguish between the original image and the 'fake' image produced by the generative model. Both the models work simultaneously until the optimal efficiency is achieved. \newline \newline
The required level of accuracy is an amalgamation of the generative and discriminative model both producing high level of accuracy. The better the generative model works, the more the 'fake' image resembles the original image. Consequently, the error rate is high in the result of the discriminative model. \newline \newline
Initially, the discriminator gives higher values to the original images which is used by the generative model to learn better and produce better quality 'fake' images. This in turn reduces the accuracy of the discriminative model as it fails to discriminate between the original and the fake image. This is when we know that both out neural networks are successfully achieving their purpose.
\newline
This is briefly explained in the methodology part later and is shown in the diagram. We finally used latent space interpolation to get the desired results.
 
\section{Initial idea}

The initial proposal presented was to train a generative adversarial network (GAN) on the MNIST dataset with its dimensionality reduced by using principle component analysis(PCA). 

For certain models features or dimensions can decrease it's accuracy due to the large amount of data that needs to be generalized, famously known as the curse of dimensionality. The reduction of dimensionality is a way to reduce the complexity of a model and to avoid overfitting, a condition when the model learns the training data too well and fails to provide useful output when exposed to new unseen data. This is where PCA comes in. 

"Principal component analysis is a mathematical algorithm that reduces the dimensionality of the data while retaining most of the variation in the data set."(Jolliffe \& Cadima, 2016)

GANs is used to generate completely new datapoints, to present photorealistic images. However it trains two networks, the generator and the discriminator, simultaneously on the images provided which proves to be very computationally expensive and in some cases overfits the data.

The objective of the project was to apply principle component analysis on our selected dataset and provide the reduced dimensionality dataset to train the GAN. This would essentially reduced the intensive operation, making it more functional. Unfortunately, we soon ran into some issues. We managed to successfully implement PCA to the data and extract the most important principal components, reducing the dimensionality in the process.

\subsection{Failure}
However, the new dataset was no longer representative of the original dataset, using this as an input to the GAN did not yield an expected result as it could not learn the features needed to generate a new image resembling the original dataset, therefore we ended up with a distorted image that we could not use and eventually decided to pivot.

This is partially due to the nature of PCA. Most of the energy is stored in the first few variables, however at least 200 variables are needed for denoising. This means often the top 10\% variables have weights x10,000 times bigger than the bottom 90\% percentile. Given that our dataset needs to be normalized for training the data started to become ambiguous. 

\begin{center}
    \includegraphics[scale=.75]{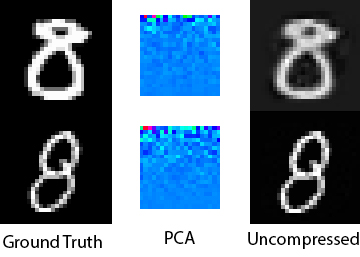}
\end{center}

As seen the PCA only holds most of its information inside of the first few layers. Another problem of variance is created.  

\begin{center}
    \includegraphics[scale=.75]{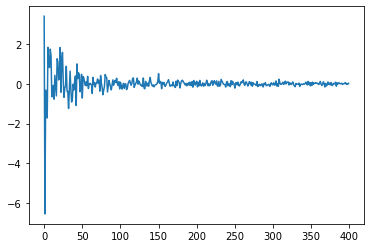}
\end{center}

This can also be seen with the PCA representation of the 8's in the top left the figure 8's are distinctive however later on they become more ambiguous and more useful for noise reduction.

\begin{center}
    \includegraphics[scale=.25]{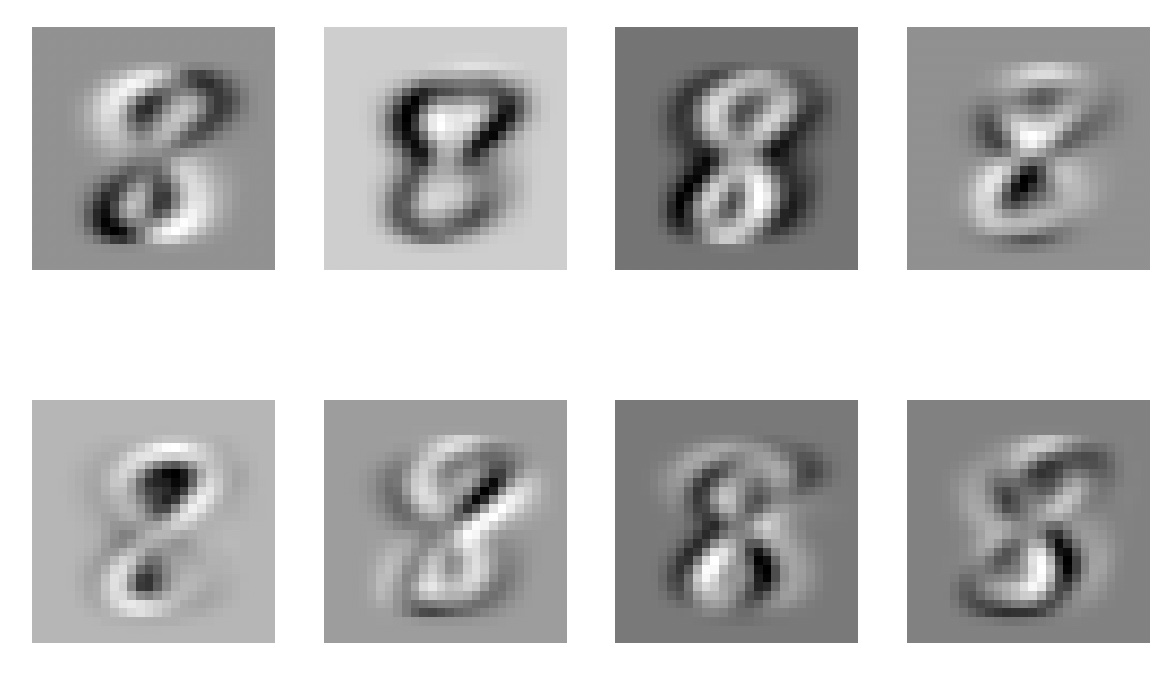}
\end{center}

\section{New idea}
Our new idea continues to work with the MNIST data set. However, we now work towards the idea of showcasing linear interpolation on latent space using a generative adversarial network. 

In its simplest form linear interpolation involves the generation of new values or points based on the existing set of values. Linear interpolation is primarily achieved by geometrically rendering a straight line between two adjacent points on a graph or plane. All points on the line connecting the two points other than the original two can be considered interpolated values. 

Using this idea we can interpolate data in the latent space and generate new data samples using our GANs model. The latent space is a representation of compressed data in which similar data points are closer together in space. This is particularly useful for learning data features, such as writing patterns in our dataset and for finding simpler representations of data for analysis.

Applying this to the context of our project would mean that through latent space interpolation along with GANs we would be able to see new handwritten numbers be generated in real-time as digits shift from one to another.

Finally, we plan on making this an interactive experience for the user by making it available through a website. Our model is trained in browser to facilitate mobility and accessibility for users and can be accessed on desktops and mobile phones alike. 

\section{Tools}
\subsection{PyTorch}
PyTorch is an open source machine learning library developed by Facebook's AI Research lab.
Its applications vary from computer vision and natural language processing to deep convolutional neural networks. It is a Python-based scientific computing package targeted to users as a deep learning research platform that provides maximum flexibility and speed.

\subsection{ONNX}
ONNX, Open Neural Network Exchange, is an open format built to represent machine learning models. ONNX defines a common set of operators used in machine and deep learning models along with a common file format to enable AI developers to use models with a variety of frameworks, tools, runtimes, and compilers. This enables us to design a deep learning model using the PyTorch framework and deploy it in a browser. 

\subsection{Generative Adversarial Networks - GANs}
Generative Adversarial Networks (GANs) are a class of neural networks that are used for unsupervised learning. It was developed and introduced by Ian J. Goodfellow in 2014. GANs consist of a system of two competing neural network models that work together to capture and analyze properties in a dataset and produce a new unseen dataset. GANs can also be used to created new datasets for supervised learning for certain classification problems. 

\begin{center}
    \includegraphics[scale=0.75]{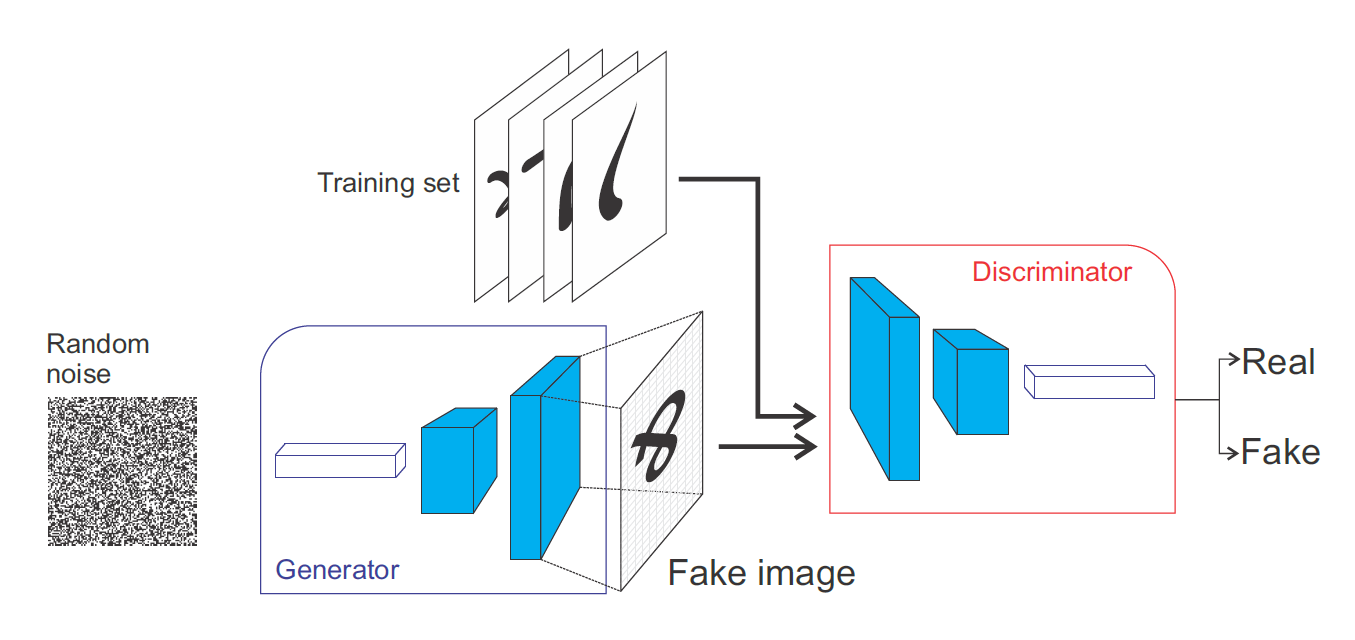}
\end{center}

Generate realistic photographs, face aging, photo blending, super resolution and even deep fakes are just some of the impressive applications of GANs.

\subsection{MNIST}
MNIST classification is a classic problem in machine learning. The problem is to look at greyscale 28x28 pixel images of handwritten digits and determine which digit the image represents, for all the digits from zero to nine.

\begin{center}
    \includegraphics[scale=0.65]{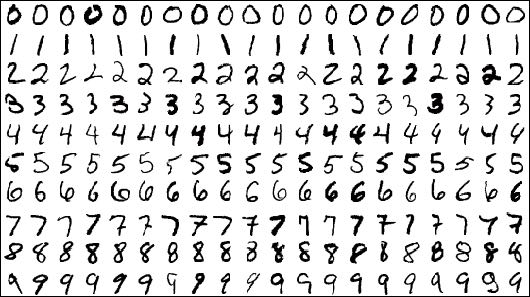}
\end{center}

The dataset is available in different testing and training sizes which are accessible \textcolor{blue}{\href{https://chromium.googlesource.com/external/github.com/tensorflow/tensorflow/+/r0.7/tensorflow/g3doc/tutorials/mnist/download/index.md}{here.}}

\section{Literature review}
 
Since Ian J. Goodfellow proposed and developed the idea of Generative Adversarial Nets i.e. GANs in 2014, there have been variety of experiments and researches involving this topic. GANs were described as a framework for training two models (i.e. the Generator G and the Discriminator) through an adversarial process. Both the models were had their own functionality with respect to the dataset. The Generator was required to’ capture’ the data distribution whereas the Discriminator was supposed to correctly discriminate between the generated points and the ones from the real dataset.  Both the models display a minimax game functionality, one model maximizes the classification error (i.e. the Generator tries to fool the Discriminator and create images that are difficult to classify) and the Discriminator tries to minimize the classification error. This results in a saddle-like graph of the functionality of both models. 

The work talks about how both the models are trained through unsupervised learning and follow back propagation and compete with each other in an adversarial modeling framework. The theoretical results are more probability based than deterministic and further algorithms are identified to apply practicality to the theory based approach.

Since our research was not just GANs based but GANs in Latent Space, the second literature piece was relevant to a great extent. Our research involved linear interpolation in latent space i.e. a space where all newly generated samples are in the form of compressed data where similar data points are closer to each other. The paper explores how success of GANs is specific to their performance in saddle point optimization and parameterizing the Generator and Discriminator as Deep Convolutional Neural Networks. The research went on to introduce Generative Latent Optimization (GLOs) which are similar in functioning to GANs but perform without the basic adversarial optimization scheme of GANs.

\section{Novelty}
Even though there has not been much work done on the latent space of GANs, we managed to successfully model it using linear interpolation and with the other tools at hand. We are proud to say that this is the first browser based interactive latent extrapolation. Users can generate new hand-written digits and see digits morph into others all with the press of a button. This is just the first roll-out, we hope to continue working and release updates to improve user experience.

\section{Methodology}
\subsection{Training GANs}

The GANs model is based on two neural networks working with each other. The neural network corresponding to the generative part of the model follows the unsupervised learning approach on the data that we provide. This is then responsible for creating new data instances. It then captures the distribution of data and maximises the final classification error.

The neural network corresponding to the discriminator part is responsible for minimizing the final classification error and discriminate between the real images and the ones generated i.e. opposing the generator neural network. The discriminator does this by using binary classification with the help of an activation function that gives a certain probability for the authenticity of the image.

In our context of the MNIST dataset we could say that the goal of the generator is to generate passable hand-written digits that are similar to the original dataset. Whereas the goal of the discriminator is to identify images coming from the generator as fake.

The GANs are formulated as a minimax game with value function V(D,G),
We train the discriminator, D, to maximize the probability of assigning the
correct label to both training examples and samples from the generator, G. We simultaneously train G to produced instances that minimize the chance of being identified by D. It can be mathematically described by the formula below:

\begin{equation}
    min_G max_D V(D,G)= E_{x_{\textasciitilde}p_{data}(x)}[log D(x)] + E{z_{\textasciitilde}p_{z}(z)}[log(1 - D(G(z)))].
\end{equation}

Our final generator is just a dense neural networks as seen below. This network has 1,486,352 parameters, we will try to reduce this later on once ONNX supports more advanced operators. 

\begin{center}
    \includegraphics[scale=0.25]{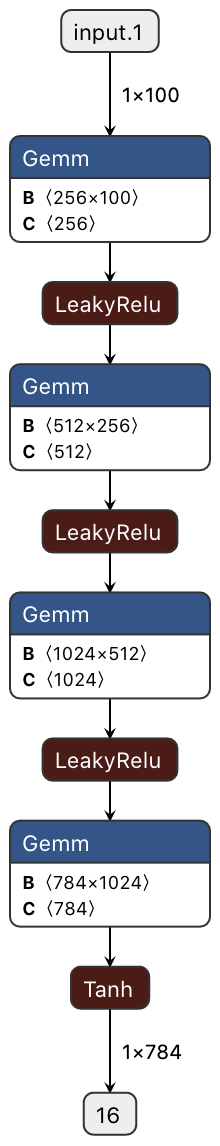}
\end{center}

\subsection{Latent Space Interpolation}

Latent space Interpolation can be thought of as a continuous walk between to points. With vectors, this can be thought of as a walk all the points to the other. If the known points are given by the vectors $S$ and $T$, the linear interpolation is the straight line between these points. This is amazing because it allows us to see in between the numbers, and now the GAN interprets similar numbers. 

\begin{equation}
    X(t)_i = (1-t)S_i + tT_i, 0\leq t \leq 1
\end{equation}

\section{Final Product}
After much revision we present the final version of our project. It is now live and can be accessed with the link to it \textcolor{blue}{\href{https://mazy1998.github.io/browserGAN/}{here}}.
Shown below is a screenshot of the user interface encountered once you arrive on the web-page.
\begin{center}
    \includegraphics[scale=0.5]{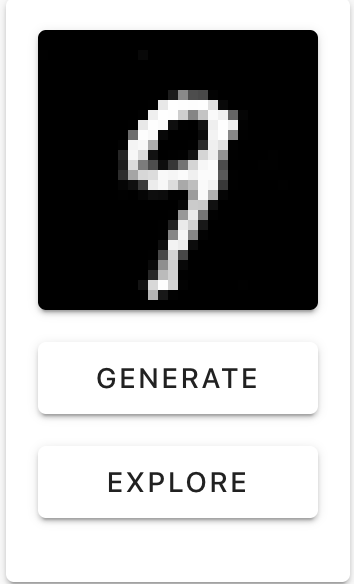}
\end{center}
Pressing the 'Generate' button will generate a single new image for a hand-written digit. Pressing the 'Explore' button begins the interpolation action and you get to see digits shift from one number to another. It must be noted that the first time the button is pressed it may take a few seconds before the result appears on screen as the model is being trained in the background and requires some time to perform all the computations needed to accomplish the task.

\section{Limitations / Challenges}
Since we were using ONNX, we were limited to only the operators it provided. Despite the many functionalities ONNX provides, it does not have an operator for $\textit{conv2dtranspose}$ therefore we had to settle for $\textit{convtranspose}$ which limits the upsampling operation in contrast to its 2d version.\\

Furthermore, even though we managed to successfully implement the model and deploy it on the web the images are not as clear as we wanted. A way to sharpen the images would be to use a deep convolutional GAN (DCGAN). These are one of the most popular adaptations of GANs. It is composed of convolutional neural networks in place of multi-layer perceptrons. Although ONNX does provide the capabilities of implementing such GANs, there are many work arounds required and much understanding of the documentation is needed. We were limited by time and could not meet due to the pandemic, therefore we settled for the success we achieved with the traditional GANs.
    
\section{Conclusion}
In conclusion, GANs are very useful for many application, one of which we have used on hand-written digits. We also provided a unique method for making this more interactive through a web-page and making it more accessible through any browser on any device.. However, there is much more that could and we look forward to in the future. 

\section{Future Work}
For the future we hope to build up on the work we have done here and continue to make such fascinating applications of machine learning more accessible to the greater public. 

We plan on implementing this again using a DCGAN for a clearer image so that we can observe the latent space interpolation in action more distinctly. However, we do not want to compromise user experience and we will deploy the model using ONNX so that it may be accessed using a browser.


\end{document}